\documentclass{article}
\usepackage{ijcai15}
\usepackage{color}

\usepackage{times}
\usepackage{multirow}
\usepackage{url}
\usepackage[titletoc]{appendix}
\usepackage{xspace}
\usepackage{enumitem}
\usepackage{color}
\usepackage{url}
\usepackage{algorithm}
\usepackage[noend]{algorithmic}
\usepackage[tight]{subfigure}
\usepackage{graphicx}
\usepackage{bm}
\usepackage{amsmath,amssymb,amsthm,xfrac}
\setitemize{noitemsep,topsep=1pt,parsep=1pt,partopsep=1pt, leftmargin=12pt}
\newtheorem{lemma}{Lemma}
\newtheorem{theorem}{Theorem}

\usepackage{caption}

\newcommand{\bB}{\bm{\beta}}

\usepackage{expl3}
\ExplSyntaxOn
\newcommand\latinabbrev[1]{
  \peek_meaning:NTF . {
    #1\@}%
  { \peek_catcode:NTF a {
      #1.\@ }%
    {#1.\@}}}
\ExplSyntaxOff

\def\eg{\latinabbrev{e.g}}

\usepackage{subfloat}

\makeatletter

\newcommand{\Rmnum}[1]{\expandafter\@slowromancap\romannumeral #1@}

\newcommand{\citet}[1]{\citeauthor{#1} [\citeyear{#1}]\xspace}

\usepackage{times}
\title{Building Hierarchies of Concepts via Crowdsourcing}
\author{
Yuyin Sun\\
University of Washington\\
\texttt{sunyuyin@cs.washington.edu} \\
\And
Adish Singla\\
ETH Zurich\\
\texttt{adish.singla@inf.ethz.ch} \\
\AND
Dieter Fox\\
University of Washington\\
\texttt{fox@cs.washington.edu} \\
\And
Andreas Krause\\
ETH Zurich\\
\texttt{krausea@ethz.ch} \\
}

\begin{document}

\maketitle

\begin{abstract}
  Hierarchies of concepts are useful in many applications from navigation to
  organization of objects. Usually, a hierarchy is created in a centralized
  manner by employing a group of domain experts, a time-consuming and expensive
  process. The experts often design one single hierarchy to best explain the
  semantic relationships among the concepts, and ignore the natural
  uncertainty that may exist in the process. In this paper, we propose a
  crowdsourcing system to build a hierarchy and furthermore capture the
  underlying uncertainty. Our system maintains a distribution over possible
  hierarchies and actively selects questions to ask using an information gain
  criterion. We evaluate our methodology on simulated data and on a set of real
  world application domains. Experimental results show that our system is robust
  to noise, efficient in picking questions, cost-effective and builds high
  quality hierarchies.
\end{abstract}

\vspace{-5mm}
\section{Introduction}
\vspace{-1.5mm}
Hierarchies of concepts and objects are useful across many real-world applications and scientific domains. Online shopping portals such as~\cite{amazon:2015} use product catalogs to organize their products into a hierarchy, aiming to simplify the task of search and navigation for their customers. Sharing the goal of organizing objects and information,  hierarchies are prevalent in many other domains such as in libraries to organize books~\cite{Dewey:1876} or web portals to organize documents by topics. Concept hierarchies also serve as a natural semantic prior over concepts, helpful in a wide range of Artificial Intelligence (AI) domains, such as natural language processing~\cite{Bloehdorn:2005} and computer vision~\cite{Deng:2009,Lai:aaai2011}.

Task-dependent hierarchies, as in product catalogs, are expensive and time-consuming to construct. They are usually built in a centralized manner by a group of domain experts. This process makes it infeasible to create separate hierarchies for each specific domain. On the other hand, in the absence of such specific hierarchies, many applications use a general-purpose pre-built hierarchy (for example, WordNet~\cite{Fellbaum:1998}) that may be too abstract or inappropriate for specific needs. An important question in this context is thus \emph{How can we cost-efficiently build task-dependent hierarchies without requiring domain experts?}

Attempts to build hierarchies using fully automatic methods~\cite{Blei:2003} have failed to capture the relationships between concepts as perceived by people. The resulting hierarchies perform poorly when deployed in real-world systems. With the recent popularity of crowdsourcing platforms, such as Amazon Mechanical Turk (AMT), efforts have been made in employing non-expert workers (the crowd) at scale and low cost, to build hierarchies guided by human knowledge. \citet{Chilton:2013} propose the \textsc{Cascade} workflow that converts the process of building a hierarchy into the task of multi-label annotation for objects. 
However, acquiring multi-label annotations for objects is expensive and might be uninformative for creating hierarchies. This leads to the question \emph{How can we actively select simple and useful questions that are most informative to the system while minimizing the cost?}

Most existing methods (including \textsc{Cascade}) as well as methods employing domain experts usually generate only a single hierarchy aiming to best explain the data or the relationships among the concepts. This ignores the natural ambiguity and uncertainty that may exist in the semantic relationships among the concepts, leading to the question \emph{How can we develop probabilistic methods that can account for this uncertainty in the process of building the hierarchy?}

{\bf Our Contributions.} In this paper, we propose a novel crowdsourcing system for inferring hierarchies of concepts,  tackling the questions posed above. We develop a principled algorithm powered by the crowd,  which is robust to noise, efficient in picking questions, cost-effective, and builds high quality hierarchies. We evaluate our proposed approach on synthetic problems, as well as on real-world domains with data collected from AMT workers, demonstrating the broad applicability of our system.


The remainder of this paper is structured as follows: After discussing related work in Section~\ref{sec:related}, we will present our method in Section~\ref{sec:approach}. We continue with experiments in Section~\ref{sec:experiments} and conclude in Section~\ref{sec:conclusion}.

\section{Related Work}
\label{sec:related}

Concept hierarchies have been helpful in solving natural language processing tasks, for example, disambiguating word sense in text retrieval~\cite{Voorhees:1993}, information extraction~\cite{Bloehdorn:2005}, and machine translation~\cite{Knight:1993}. Meanwhile, hierarchies between object classes have been deployed in the computer vision community to improve object categorization with thousands of classes and limited training images~\cite{Rohrbach:2011}, scalable image classification~\cite{Deng:2009,Deng:2013,Lai:aaai2011}, and image annotation efficiency~\cite{Deng:2014}. In these methods, it is usually assumed that the hierarchies have already been built, and the quality of the hierarchies can influence the performance of these methods significantly.

The traditional way of hierarchy creation is to hire a small group of experts to build the hierarchy in a centralized manner~\cite{Fellbaum:1998}, which is expensive and time consuming. Therefore, people develop automatic or semi-automatic methods to build hierarchies. For instance, vision based methods, such as, \citet{Sivic:2008} and \citet{Bart:2008} build an object hierarchy using visual feature similarities. However, visually similar concepts are not necessarily similar in semantics.

Another type of methods for hierarchy creation is related to ontology learning from text and the web~\cite{Buitelaar:2005,Wong:2012,Carlson:2010}. The goal of ontology learning is to extract terms and relationships between these concepts. However, the focus of these techniques is on coverage, rather than accuracy, and the hierarchies that can be extracted from these approaches are typically not very accurate.  Since the taxonomy is the most important relationship among ontologies, many works have been focusing on building taxonomy hierarchies. For example, co-occurrence based methods~\cite{Budanitsky:1999} use word co-occurrence to define the similarity between words, and build hierarchies using clustering. These methods usually do not perform well because they lack in common sense~\cite{Wong:2012}. On the other hand, template-based methods~\cite{Hippisley:2005} deploy domain knowledge and can achieve higher accuracy. Yet, it is hard to adapt template-based methods to new domains. Knowing the fact that humans are good at common sense, and domain adaptation, involvement of humans in hierarchy learning remains highly necessary and desirable~\cite{Wong:2012}.


The popularity of crowdsourcing platforms has made cheap human resources available for building hierarchies. For example, \textsc{Cascade}~\cite{Chilton:2013} uses multi-label annotations for items, and deploys label co-occurrence to generate a hierarchy. \textsc{Deluge}~\cite{Bragg:2013} improves the multi-label annotation step in~\textsc{Cascade} using decision theory and machine learning to reduce the labeling effort. However, for both pipelines, co-occurrence of labels does not necessarily imply a connection in the hierarchy. Furthermore, both methods can build only a single hierarchy, not considering the uncertainty naturally existing in hierarchies.

Orthogonal to building hierarchies, \citet{Mortensen:2006} use crowdsourcing to verify an existing ontology. Their empirical results demonstrate that non-expert workers are able to verify structures within a hierarchy built by domain experts. Inspired by their insights, it is possible to gather information of the hierarchy structure by asking simple true-or-false questions about the ``ascendant-descendant'' relationship between two concepts. In this work, we propose a novel method of hierarchy creation based on asking such questions, and fusing the information together.

\section{Approach}
\label{sec:approach}


The goal of our approach is to learn a hierarchy over a domain of concepts,
using input from non-expert crowdsourcing workers.  Estimating hierarchies
through crowdsourcing is challenging, since answers given by workers are
inherently noisy, and, even if every worker gives her/his best possible answer,
concept relationships might be ambiguous and there might not exist a single
hierarchy that consistently explains all the workers' answers.  We deal with
these problems by using a Bayesian framework to estimate probability
distributions over hierarchies, rather than determining a single, best
guess. This allows our approach to represent uncertainty due to noisy, missing,
and possibly inconsistent information.  Our system interacts with crowdsourcing
workers iteratively while estimating the distribution over hierarchies. At each
iteration, the system picks a question related to the relationship between two
concepts in the hierarchy, presents it to multiple workers on a crowdsourcing
platform, and then uses the answers to update the distribution. The system keeps
asking questions until a stopping criterion is reached. In this work we set a
threshold for the number of asked questions.

\subsection{Modelling Distributions over Hierarchies}

The key challenge for estimating distributions over hierarchies is the huge number of possible hierarchies, or trees, making it intractable to directly represent the distribution as a multinomial. Consider the number of possible hierarchies of $N$ concepts is $(N+1)^{N-1}$ (we add a fixed root node to the concept set), which results in 1,296 trees for 5 concepts, but already $2.3579e+09$ trees for only 10 concepts.  We will now describe how to represent and estimate distributions over such a large number of trees.

Assume that there are $N$ concept nodes indexed from $1$ to $N$, a fixed root node indexed by $0$, and a set of possible directed edges $\mathcal{E} = \{e_{0,1}, \ldots,e_{i,j}, \ldots, e_{N,N}\}$ indexed by $(i,j)$, where $i\neq j$. A hierarchy $T \subset \mathcal{E}$ is a set of $N$ edges, which form a valid tree rooted at the $0$-th node (we use the terms hierarchy and tree interchangeably). All valid hierarchies form the sample space $\mathcal{T} = \{T_1, \ldots, T_M\}$. The prior distribution $\pi^{0}$ over $\mathcal{T}$ is set to be the uniform distribution. 


Due to the intractable number of trees, we use a compact model to represent
distributions over trees:
\begin{equation}\label{eq:tree_distribution}
P(T| W) = \frac{\prod_{e_{i,j}\in T} W_{i,j}}{Z(W)}\,,
\vspace{-1ex}
\end{equation}
\looseness -1 where $W_{i,j}$ is a non-negative weight for the edge $e_{i,j}$, and $Z(W) =
\sum_{T'\in \mathcal{T}} \prod_{e_{i,j}\in T'}W_{i,j}$ is the partition function. Given $W$, inference is very efficient. For example, $Z(W)$ can be analytically computed, utilizing the Matrix Theorem~\cite{Tutte:1984}. This way, we can also analytically compute marginal probabilities over the edges, i.e., $P(e_{i,j})$. 
The tree with the highest probability can be found via 
the famous algorithm of~\citet{Chu:1965}. A uniform prior is incorporated by initially setting all $W_{i,j}$ to be the same positive value.


Our system maintains a posterior distribution over hierarchies.
Given a sequence of questions regarding the
structure of the target hierarchy $q^{1}, \ldots, q^{(t)}, \ldots$ , along with the answers $a^{1}, \ldots, a^{(t)},
\ldots$ 
the posterior
$P(T|W^{(t)})$ at time $t$  is obtained by Bayesian inference
\begin{equation}
P(T|W^{(t)}) \propto P(T|W^{(t-1)})f(a^{(t)}|T).\label{eq:bayesian_update}
\vspace{-0.4em}
\end{equation}
Hereby, $f(a^{(t)}|T)$ is the likelihood of obtaining answer $a^{(t)}$ given a
tree $T$, specified below to simplify the notation. 

So far, we have not made any assumptions about the form of questions asked. Since
our system works with non-expert workers, the questions
should be as simple as possible.  As discussed above, we resort to questions
that only specify the relationship between pairs of concepts. We will discuss
different options in the following sections.

\subsection{Edge Questions}

Since a hierarchy is specified by a set of edges, one way to ask questions could
be to ask workers about immediate parent-child relationships between concepts,
which we call {\em edge questions}.  Answers to edge questions are highly informative,
since they provide direct information about whether there is an edge between
two concepts in the target hierarchy.

Let $e_{i,j}$ denote the question of whether there is an edge between node $i$
and $j$, and $a_{i,j}\in \{0, 1\}$ denote the answer for $e_{i,j}$. $a_{i,j} =
1$ indicates a worker believes there is an edge from node $i$ to $j$, otherwise
there is no edge.  The likelihood function for edge questions is defined as
follows:
\begin{align}\label{eq:likelihood_edge}
f(a_{i,j}|T) = \begin{cases}
(1-\gamma)^{a_{i,j}}\gamma^{1-a_{i,j}}\mbox{, if }e_{i,j}\in T\\
\gamma^{a_{i,j}}(1-\gamma)^{1-a_{i,j}}\mbox{, otherwise}
\end{cases},
\end{align}
where $\gamma$ is the noise rate for wrong
answers. Substituting~(\ref{eq:likelihood_edge}) into~(\ref{eq:bayesian_update})
leads to an analytic form to update edge weights: 
%
%
%

\begin{align}
\vspace{-1ex}
W_{i',j'}^{(t)} = \begin{cases}
W_{i',j'}^{(t-1)}(\frac{1-\gamma}{\gamma})^{(2a_{i,j} - 1)}\mbox{, if } i'=i \land j'=j\\
W_{i',j'}^{(t-1)}\mbox{, otherwise}
\end{cases}.
\end{align}

An edge question will only affect weights for that edge. 
Unfortunately, correctly answering such questions is difficult and requires global knowledge of
the complete set (and granularity) of concepts.  For instance, while the
statement ``Is \emph{orange} a direct child of \emph{fruit} in a food item
hierarchy?'' might be true for some concept sets, it is not correct in a
hierarchy that also contains the more specific concept \emph{citrus fruit},
since it separates \emph{orange} from \emph{fruit} (see also
Fig.~\ref{fig:hierarchies}).


\subsection{Path Questions}

To avoid the shortcomings of edge questions, our system resorts to asking less
informative questions relating to general, ascendant-descendant relationships
between concepts.  These {\em path questions} only provide information about the
existence of directed paths between two concepts and are thus, crucially,
independent of the set of available concepts.  For instance, the path question
``Is \emph{orange} a type of \emph{fruit}?''  is true independent of the
existence of the concept \emph{citrus fruit}.  While such path questions are
easier to answer, they are more challenging to use when estimating the
distribution over hierarchies.

To see, let $p_{i,j}$ denote a path question and $a_{i,j}\in\{0, 1\}$ be the
answer for $p_{i,j}$. $a_{i,j}=1$ indicates a worker believes there is a path
from node $i$ to $j$. The likelihood function is
\begin{align}\label{eq:likelihood_path}
f(a_{i,j}|T) = \begin{cases}
(1-\gamma)^{a_{i,j}}\gamma^{1-a_{i,j}}\mbox{, if }p_{i,j}\in T\\
\gamma^{a_{i,j}}(1-\gamma)^{1-a_{i,j}}\mbox{, otherwise}
\end{cases},
\end{align}
where $p_{i,j}\in T$ simply checks whether the path $p_{i,j}$ is contained in
the tree $T$.

Unfortunately, the likelihood function for path questions is not conjugate of
the prior. Therefore, there is no analytic form to update weights. Instead, we
update the weight matrix by performing approximate inference. To be more
specific, we find a $W^*$ by minimizing the KL-divergence~\cite{Kullback:1951}
between $P(T|W^*)$ and the true posterior:
\begin{equation}
W^* = \arg\min_{W} KL(P(T|W^{(t)})\|P(T|W))
\vspace{-1.2ex}
\end{equation}

It can be shown that minimizing the KL-divergence can be achieved by minimizing
the following loss function
%
\begin{equation}
L(W) = -\sum_{T\in \mathcal{T}} P(T|W^{(t)})\log P(T|W).\label{eq:ori_neg_log_likelihood}
\vspace{-1ex}
\end{equation} 



Directly computing~(\ref{eq:ori_neg_log_likelihood}) involves enumerating all trees in $\mathcal{T}$, and is therefore intractable. Instead, we use a Monte Carlo method to estimate~(\ref{eq:ori_neg_log_likelihood}), and minimize the estimated loss to update $W$. To be more specific, we will generate i.i.d.~samples $\mathcal{\widetilde{T}}
=(T_1, \ldots, T_m)$
from $P(T|W^{(t)})$, which defines an empirical distribution $\tilde{\pi}^{(t)}$ of the samples,
%
with estimated loss
\begin{equation}\label{eq:estimate_neg_log_loss}
L_{\tilde{\pi}}(W) = -\sum_{T\in \mathcal{\widetilde{T}}} \tilde{\pi}(T) \log P(T|W),
\vspace{-1ex}
\end{equation}
the negative log-likelihood of $P(T|W)$ under the samples.

\subsubsection{Sampling Hierarchies from the Posterior} 

If a weight matrix $W$ is given, sampling hierarchies from the distribution
defined in~(\ref{eq:tree_distribution}) can be achieved efficiently, for
example, using a loop-avoiding random walk on a graph with $W$ as the adjacency
matrix~\cite{Wilson:1996}. Therefore, we can sample hierarchies from the prior
$P(T|W^{(t-1)})$. Noticing that the posterior defined
in~(\ref{eq:bayesian_update}) is a weighted version of $P(T|W^{(t-1)})$, we can
generate samples for the empirical distribution $\tilde{\pi}$ via importance
sampling, that is, by reweighing the samples from $P(T|W^{(t-1)})$ with the
likelihood function $f(a^{(t)}|T)$ as importance weights.

%

%



\subsubsection{Regularization}
%
%

Since we only get samples from $P(T|W^{(t)})$, the estimate of~(\ref{eq:estimate_neg_log_loss}) can be inaccurate. To avoid overfitting to the sample, we add an $\ell_1$-regularization term to the objective function. We also optimize $\Lambda = \log W$ rather than $W$ so as to simplify notation. The final objective is as follows:

\begin{equation}\label{eq:loss_function}
L_{\tilde{\pi}}^{\bm{\beta}}(\Lambda^{(t)}) \!=\! -\!\sum_{T\in \mathcal{T}} \tilde{\pi}^{(t)}(T)\log P(T|\Lambda^{(t)}) + \sum_{i,j} \beta|\lambda_{i, j}|.
\vspace{-0.4ex}
\end{equation}

%

\subsubsection{Optimization Algorithm}
\label{sec:algorithm}

We iteratively adjust $\Lambda$ to minimize~(\ref{eq:loss_function}). At each iteration, the algorithm adds $\Delta$ to the original $\Lambda$, resulting in $\Lambda' = \Lambda + \Delta$. We optimize $\Delta$ to minimize an upper bound on the change in $L^{\beta}_{\tilde{\pi}}$, given by 
\begin{align}
L^{\bB}_{\tilde{\pi}}(\Lambda') \!-\! L^{\bB}_{\tilde{\pi}}(\Lambda)\nonumber
\!\leq\! & \sum_{i,j} [-\delta_{i,j}\tilde{P}(e_{i,j}) + \frac{1}{N}P(e_{i,j}|\Lambda)(e^{N\delta_{i,j}} - 1) \nonumber\\
&+ \beta(|\lambda_{i,j}'| - |\lambda_{i,j}|)] + C,\label{eq:upper_bound}
\end{align}
where $\tilde{P}(e_{i,j}) = \sum_{T\in \mathcal{T}: e_{i,j}\in T} \tilde{\pi}(T)$ is the empirical marginal probability of $e_{i,j}$, and $C$ is a constant w.r.t.~$\delta_{i,j}$. The derivation is presented in~\cite{Sun:2015}.


%
%

Minimizing the upper bound in~(\ref{eq:upper_bound}) can be done by analysing the sign of $\lambda_{i,j} + \delta_{i,j}$. By some calculus, it can be seen that the $\delta_{i,j}$ minimizing~(\ref{eq:upper_bound}) must occur when $\delta_{i,j} = -\lambda_{i,j}$, or when $\delta_{i,j}$ is either
\begin{align*}
&\frac{1}{N}\log \frac{(P(e_{i,j}) - \beta)}{P(e_{i,j}|\Lambda)},\mbox{ if } \lambda_{i,j} + \delta_{i,j} \geq 0, \text{or}\\
&\frac{1}{N}\log \frac{(P(e_{i,j}) + \beta)}{P(e_{i,j}|\Lambda)}, \mbox{ if }\lambda_{i,j} + \delta_{i,j} \leq 0.
\end{align*}
This leaves three choices for each $\delta_{i,j}$ -- we try out each and pick the one leading to the best bound. This can be done independently per $\delta_{i,j}$ since the objective in~\eqref{eq:upper_bound} is separable.
 The full algorithm to optimize $\Lambda$ based on a query answer is given in Algorithm~\ref{alg:update_weights_path}.

\setlength{\textfloatsep}{5pt}
\begin{algorithm}[tb]
   \caption{Weight Updating Algorithm}
   \label{alg:update_weights_path}
\begin{algorithmic}
   \STATE {\bfseries Input:} $W^{(t-1)}$, an answer $a^{(t)}$, $thr$ for stopping criterion
   \STATE Non-negative regularization parameters $\beta$
   \STATE {\bfseries Output:} $W^{(t)}$ that minimizes (\ref{eq:loss_function})
   \STATE Generate samples $T_1', \ldots, T_m'$ from $P(T|W^{(t-1)})$
   \STATE Use importance sampling to get empirical distribution $\tilde{\pi}$
   \STATE Initialize $\Lambda^{0} = \bm{0}$, $l = 1$.   
   \REPEAT
   \STATE For each $(i,j)$, set $\delta_{i,j} = \arg\min $ (\ref{eq:upper_bound});
   \STATE Update $\Lambda^{(l)} = \Lambda^{(l-1)} + \Delta$;
   \STATE $l = l + 1$;
   \UNTIL{$|\Delta| \leq thr$}   
   \STATE return $W^{(t)} = \exp(\Lambda^{(l)})$
\end{algorithmic}
\end{algorithm}

\subsubsection{Theoretical Guarantee}

Even though we only minimize a sequence of upper bounds, we prove 
that (see~\cite{Sun:2015}) Algorithm~\ref{alg:update_weights_path} in fact convergences to the true maximum likelihood solution:
\begin{theorem}
Assume $\beta$ is strictly positive. Then  Algorithm~\ref{alg:update_weights_path} produces a sequences $\Lambda^{(1)}, \Lambda^{(2)}, \ldots$ such that
\[\lim_{\ell \rightarrow \infty} L_{\tilde{\pi}}^{\beta}(\Lambda^{(\ell)}) = \min_{\Lambda} L_{\tilde{\pi}}^{\beta}(\Lambda).\]
\end{theorem}

%

Let $\hat{\Lambda}$ be the solution of Algorithm~\ref{alg:update_weights_path}.  We next show that if we generate enough samples $m$, the loss of the estimate $\hat{\Lambda}$ under the true posterior will not be much higher than that obtained by any distribution of the form \eqref{eq:tree_distribution}.

\begin{theorem}\label{theorem:loss_bound}
Suppose $m$ samples $\tilde{\pi}$ are obtained from any tree distribution $\pi$.
Let $\hat{\Lambda}$ minimize the regularized log loss $L_{\tilde{\pi}}^{{\beta}}(\Lambda)$ with $\beta = \sqrt{\log(N/\delta)/(m)}$. Then for every $\Lambda$ 
it holds with probability at least $1-\delta$ that
\[L_{\pi}(\hat{\Lambda}) \leq L_{\pi}(\Lambda) + 2\|\Lambda\|_1\sqrt{\log(N/\delta)/m}\]
\end{theorem}



Theorem~\ref{theorem:loss_bound} shows that the difference in performance between the density estimate computed by minimizing w.r.t.~$L_{\tilde{\pi}}^{\beta}$ and w.r.t.~the best approximation to the true posterior becomes small rapidly as the number of samples $m$ increases.


\subsection{Active Query Selection}

At each interaction with workers, the system needs to pick a question to
ask. The naive approach would be to pick questions randomly. However random
questions are usually not very informative since they mostly get \textit{No}
answers, while \textit{Yes} answers are more informative about the structure.
Instead, we propose to select the question $p^{*}$ that maximizes information
gain over the current distribution $\pi^{(t)}$, i.e.
\begin{equation}
p^{*} = \arg\max_{p_{i, j}}\max(H(\tilde{\pi}^{(t+1)}_{p_{i, j}, 1}), H(\tilde{\pi}^{(t+1)}_{p_{i, j}, 0}))\label{eq:active_utility}
\vspace{-1ex}
\end{equation}
where $H(\cdot)$ is the entropy and $\tilde{\pi}^{(t+1)}_{p_{i, j}, a_{i, j}}$
is the posterior distribution over trees after knowing the answer for $p_{i, j}$
as $a_{i, j}$.  Note that this criterion chooses the question with the highest
information gain using the \emph{less} informative answer, which we found to be
more robust than using the expectation over answers.  Since we cannot compute
the entropy exactly due to the size of the sampling space, we reuse the trees
from the empirical distribution $\tilde{\pi}$ to estimate information gain.

%
%


\subsection{Adding New Nodes to the Hierarchy}

New concepts will sometimes be introduced to a domain. In this case, how should
the new concept be inserted into an existing hierarchy? A wasteful way would be
to re-build the whole hierarchy distribution from scratch using the pipeline
described in the previous sections.  Alternatively, one might consider adding a
row and column to the current weight matrix and initializing all new entries
with the same value. Unfortunately, uniform weights do not result in uniform
edge probabilities and do thus not correctly represent the uninformed prior over
the new node's location. 

We instead propose a method to estimate the weight matrix $W$ that correctly
reflects uncertainty over the current tree and the location of the new node.
After building a hierarchy for $N+1$ nodes, the learned weight matrix is denoted
as $W_N$. We can sample a sequence of trees $(T_1, \ldots, T_m)$ from the
distribution $P(T|W_N)$. Now we want to insert a new node into the
hierarchy. Since there is no prior information about the position of the new
node, we generate a new set of trees by inserting this node into any location in
each tree in $(T_1, \ldots, T_m)$.  This new set represents a sample
distribution that preserves the information of the previous distribution while
not assuming anything about the new node. The weight matrix $W^{(0)}_{N+1}$ that
minimizes KL-divergence to this sample can then be estimated using the method
described in Section~\ref{sec:algorithm}.

\section{Experiments}
\label{sec:experiments}
In our experiments, we evaluate four aspects of our approach: (1) the performance of approximate inference to estimate the weights representing distributions over trees; (2) the efficiency of active vs.~random query strategies in selecting questions; (3) comparison to existing work; and (4) the ability to build hierarchies for diverse application domains.

\subsection{Sample-based Weight Estimation}

\begin{figure}[t]
\begin{center}
\centerline{\includegraphics[width=0.65\columnwidth]{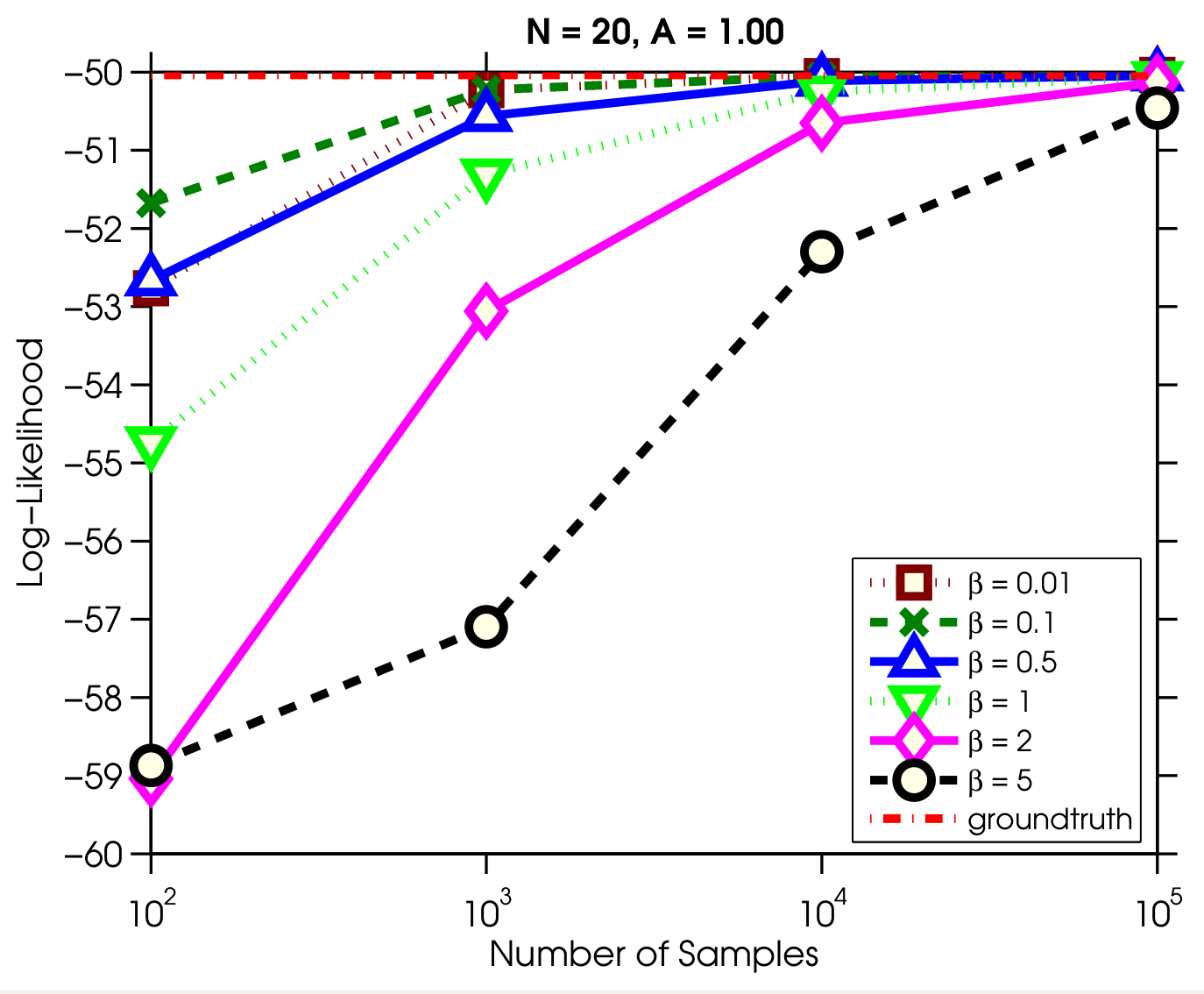}}
\caption{\small{Weight estimation performance for hierarchies with 20 nodes. X-axis is the number of samples given to the algorithm, and $\beta$ is the regularization coefficient.}}
\label{fig:weight_update}
\end{center}
\end{figure}


To evaluate the ability of Algorithm~\ref{alg:update_weights_path} to estimate a weight matrix based on samples generated from a distribution over trees we proceeded as follows. We first sample a ``ground truth'' weight matrix $W$, then sample trees according to that weight matrix, followed by estimating the weight matrix $W^{*}$ using Algorithm~\ref{alg:update_weights_path}, and finally evaluate the quality of the estimated matrix. To do so, we sample an additional test set of trees from $P(T|W)$ and compute the log-likelihood of these trees given $W^*$, where $P(T|W^*)$ is defined in~(\ref{eq:tree_distribution}). For calibration purpose, we also compute the log-likelihood of these trees under the ground truth specified by $P(T|W)$.

%

Fig.~\ref{fig:weight_update} shows the performance for $N=20$ nodes, using different values for the regularization coefficient $\beta$ and sample size $m$.  Each data point is an average over 100 runs on different weights sampled from a Dirichlet distribution (to give an intuition for the complexity of the investigated distributions, when sampling 1 Million trees according to one of the weights, we typically get about $900,000$ distinct trees).  The red line on top is the ground truth log-likelihood. As can be seen, the algorithm always converges to the ground truth as the number of samples increases. With an appropriate setting of $\beta$, the proposed method requires about $10,000$ samples to achieve a log-likelihood that is close to the ground truth. We also tested different tree sizes $N$, and got quite similar performance. Overall, we found that $\beta = 0.01$ works robustly across different $N$ and use that value in all the following experiments.


\subsection{Active vs.~Random Queries}
\label{sec:active_random}

\begin{figure*}[t]
\begin{center}
{\includegraphics[height=30ex,trim=55mm 60mm 60mm 60mm]{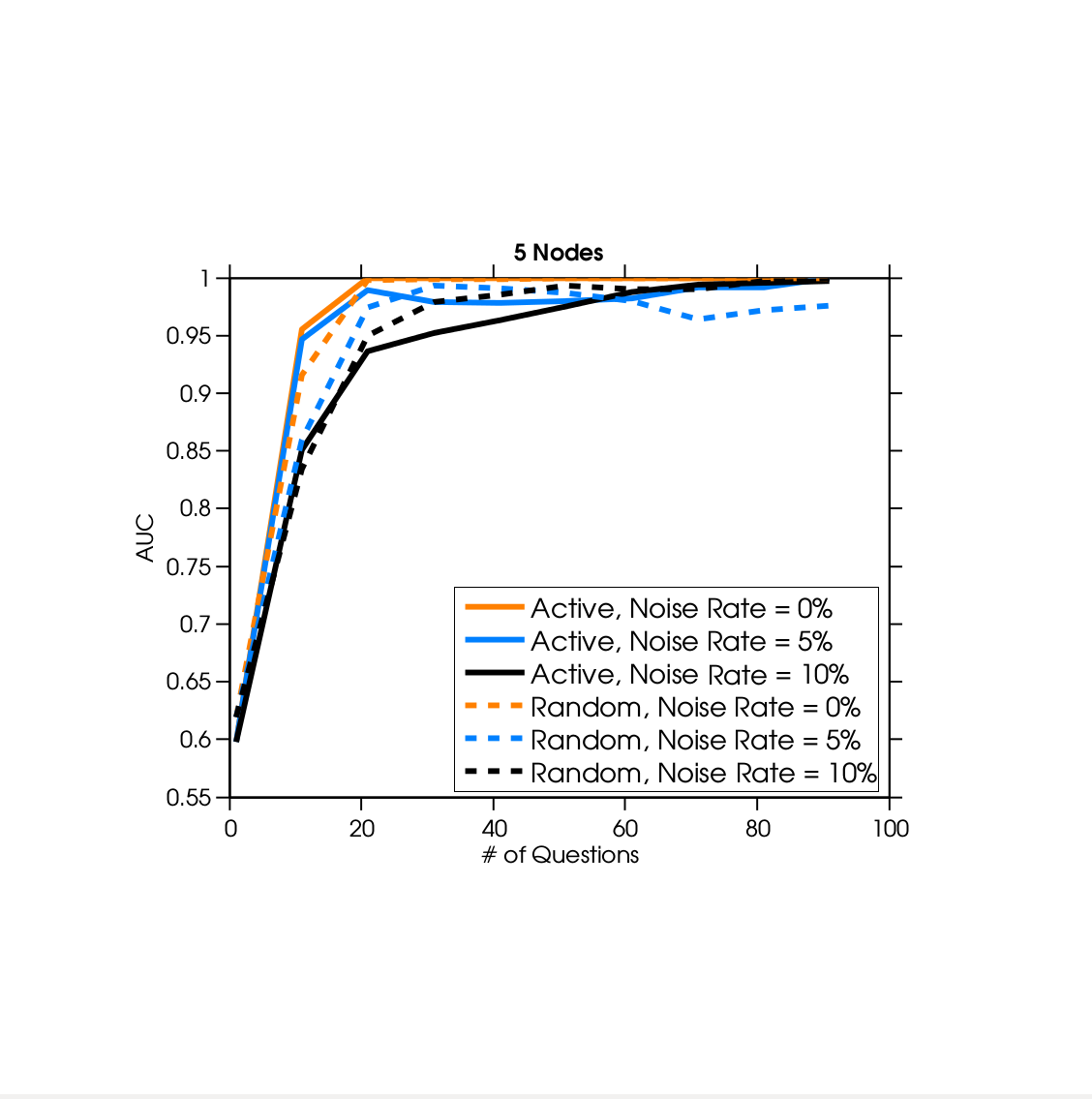}}\hfill
{\includegraphics[height=30ex,trim=55mm 60mm 60mm 60mm]{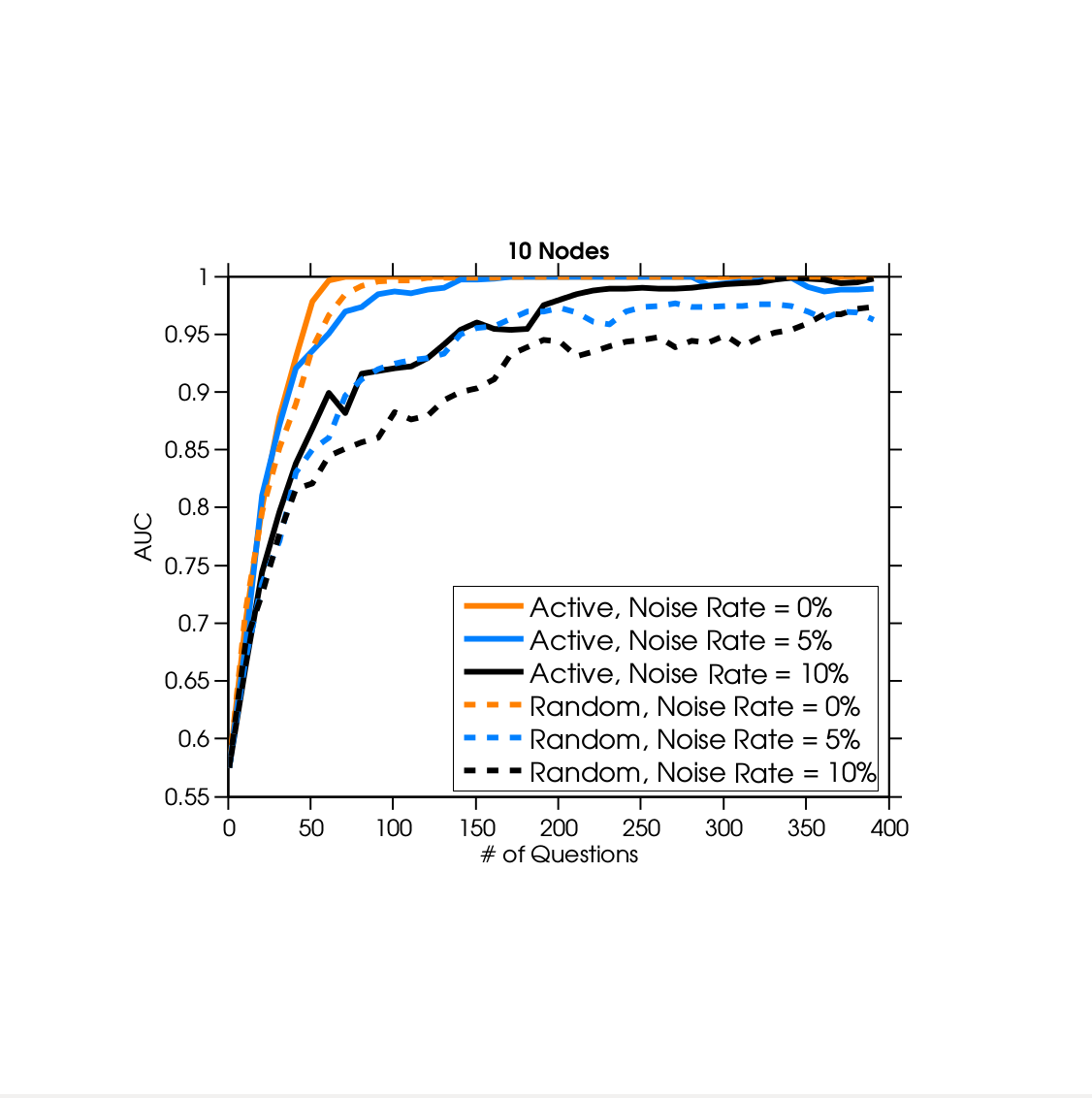}}\hfill
{\includegraphics[height=30ex,trim=55mm 60mm 60mm 60mm]{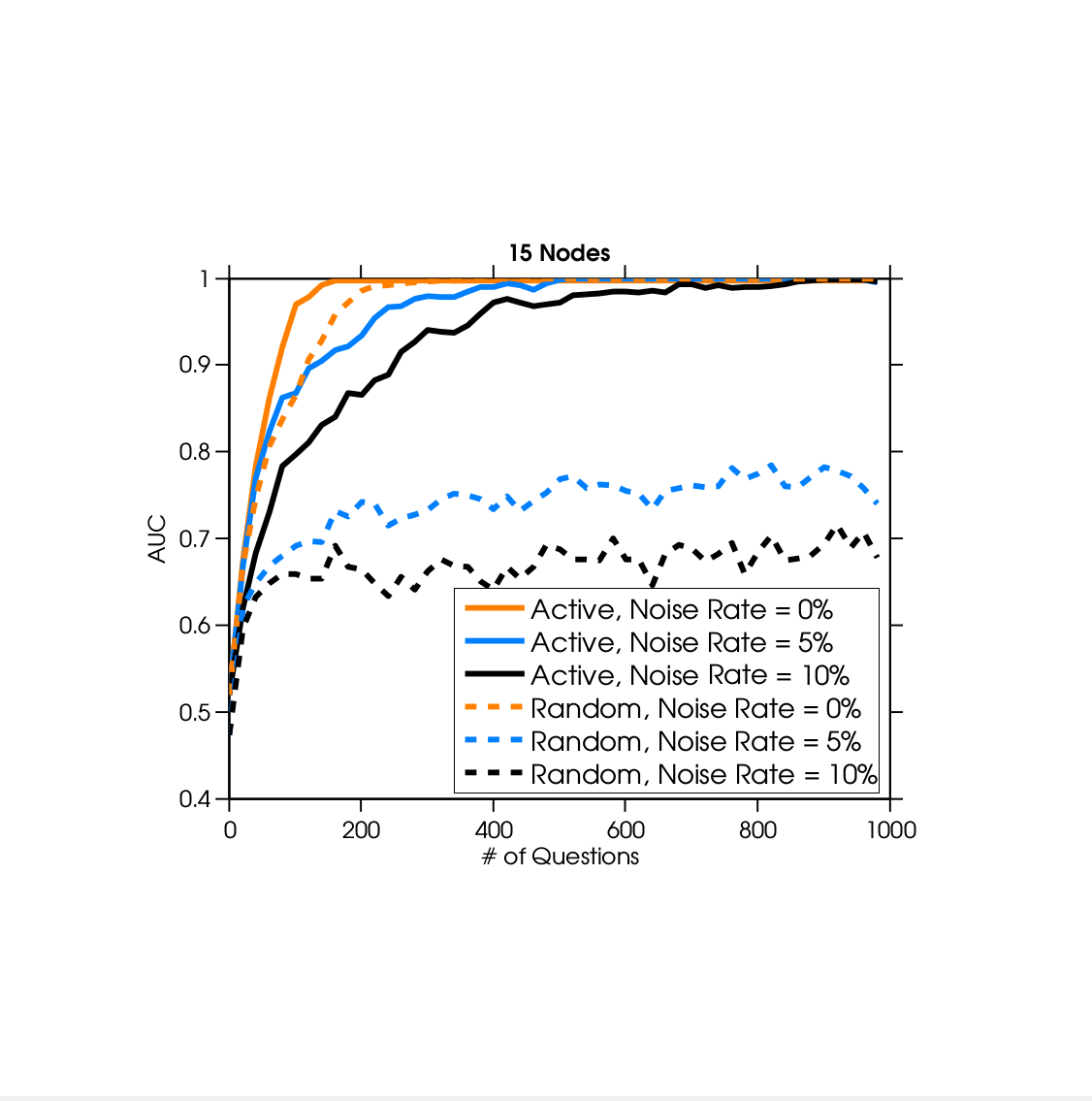}}
\caption{\small Experimental results comparing active query (solid lines) and random query (dashed lines) strategies for tree sizes ranging from (left) 5 nodes to (right) 15 nodes, using three different noise rates for answers to questions.}\label{fig:active_random}
\end{center}
\end{figure*} 


To evaluate the ability of our technique to recover the correct hierarchy, we artificially generate trees and test the performance for different tree sizes, different noise rates for path queries, and active vs.~random path queries.  To simulate a worker's response to a query, we first check whether the query path is part of the ground truth tree, and then flip the answer randomly using a pre-set noise rate $\gamma$.  To evaluate how well our approach estimates the ground truth tree, we use the marginal likelihood of the tree edges and compute the Area Under the Curve (AUC) using different thresholds on the likelihood to predict the existence or absence of an edge in the ground truth tree. The marginal likelihood of an edge, $P(e_{i,j}| W) = \sum_{T\in \mathcal{T}, e_{i,j}\in T} P(T| W)$, can be computed in closed form based on the conclusion of the Matrix Theorem~\cite{Tutte:1984}.  We also tested different evaluation measures, such as the similarity between the MAP tree and the ground truth tree, and found them to all behave similarly.


Different sized trees of \{5, 10, 15\} nodes are tested to see how the method performs as the problem gets larger. We also test different noise rates, including $0\%$, $5\%$, and $10\%$ to verify the robustness of the method. The number of samples for updating the weight matrix is fixed to $10,000$ across all experiments. For each setting, 10 different random ground truth trees are generated. The average results are reported in Fig.~\ref{fig:active_random}. The X-axis is the number of questions asked, and the Y-axis is the $AUC$. $AUC = 1$ means that all edges of the ground truth tree have higher probabilities than any other edges according to the estimated distribution over trees.

As can be seen in the figures, active queries always recover the hierarchy more efficiently than their random counterparts (random queries are generated by randomly choosing a pair of nodes).  If there is no noise in the answers, our approach always recovers the ground truth hierarchy, despite the sample-based weight update. Note that, in the noise-free setting, the exact hierarchy can be perfectly recovered by querying all $N^2$ pairs of concepts.  While the random strategy typically requires about twice that number to recover the hierarchy, our proposed active query strategy always recovers the ground truth tree using less than $N^2$ samples. 

As $N$ gets larger, the difference between active and random queries becomes more significant. While our active strategy always recovers the ground truth tree, the random query strategy does not converge for trees of size 15 if the answers are noisy. This is due to insufficient samples when updating the weights, and because random queries are frequently answered with \textit{No}, which provides little information.  The active approach, on the other hand, uses the current tree distribution to determine the most informative query and generates more peaked distributions over trees, which can be estimated more robustly with our sampling technique.  As an indication of this, for trees of size 15 and noise-free answers, 141 out of the 200 first active queries are answered with ``yes'', while this is the case for only 54 random queries.



\subsection{Comparison to Existing Work}

We compare our method with the most relevant systems \textsc{Deluge}~\cite{Bragg:2013} and \textsc{Cascade}~\cite{Chilton:2013}, which also use crowdsourcing to build hierarchies. \textsc{Cascade} builds hierarchies based on multi-label categorization, and \textsc{Deluge} improves the multi-label classification performance of \textsc{Cascade}. We will thus compare to \textsc{Deluge}, using their evaluation dataset~\cite{Bragg:2013}. This dataset has 33 labels that are part of the fine-grained entity tags~\cite{Ling:2012}. The WordNet hierarchy is used as the ground truth hierarchy.

\textsc{Deluge} queries many items for each label from a knowledge base, randomly selects a subset of 100 items, labels items with multiple labels using crowdsourcing, and then builds a hierarchy using the label co-occurrence. To classify items into multi-labels, it asks workers to vote for questions, which are binary judgements about whether an item belongs to a category. \textsc{Deluge} does active query selection based on the information gain, and considers the label correlation to aggregate the votes and build a hierarchy. We use the code and parameter settings provided by the authors of \textsc{DELUGE}.

We compare the performance of our method to \textsc{Deluge} using different amounts of votes. We compare the following settings: 1) Both methods use 1,600 votes; 2) \textsc{Deluge} uses 49,500 votes and our method uses 6,000 votes. For the first setting, we pick 1,600 votes for both, as suggested by the authors because \textsc{Deluge}'s performance saturates after that many votes. In the second setting, we compute the results of using all the votes collected in the dataset to see the best performance of \textsc{Deluge}. We choose 6,000 votes for our method because its performance becomes flat after that.

We compare both methods using AUC as the evaluation criterion. Using 1,600 votes, our method achieves a value of $0.82$, which is slightly better than \textsc{Deluge} with an AUC of $0.79$.  However, \textsc{Deluge} does not improve significantly beyond that point, reaching an AUC of $0.82$ after 49,500 votes.  Our approach, on the other hand, keeps on improving its accuracy and reaches an AUC of $0.97$ after only 6,000 queries. This indicates that, using our approach, non-expert workers can achieve performance very close to that achieved by experts (AUC = 1).  Furthermore, in contrast to our method, \textsc{Deluge} does not represent uncertainty over hierarchies and requires items for each label.

\subsection{Real World Applications}

\begin{figure*}[t]
\begin{center}
\fbox{\includegraphics[height=29ex]{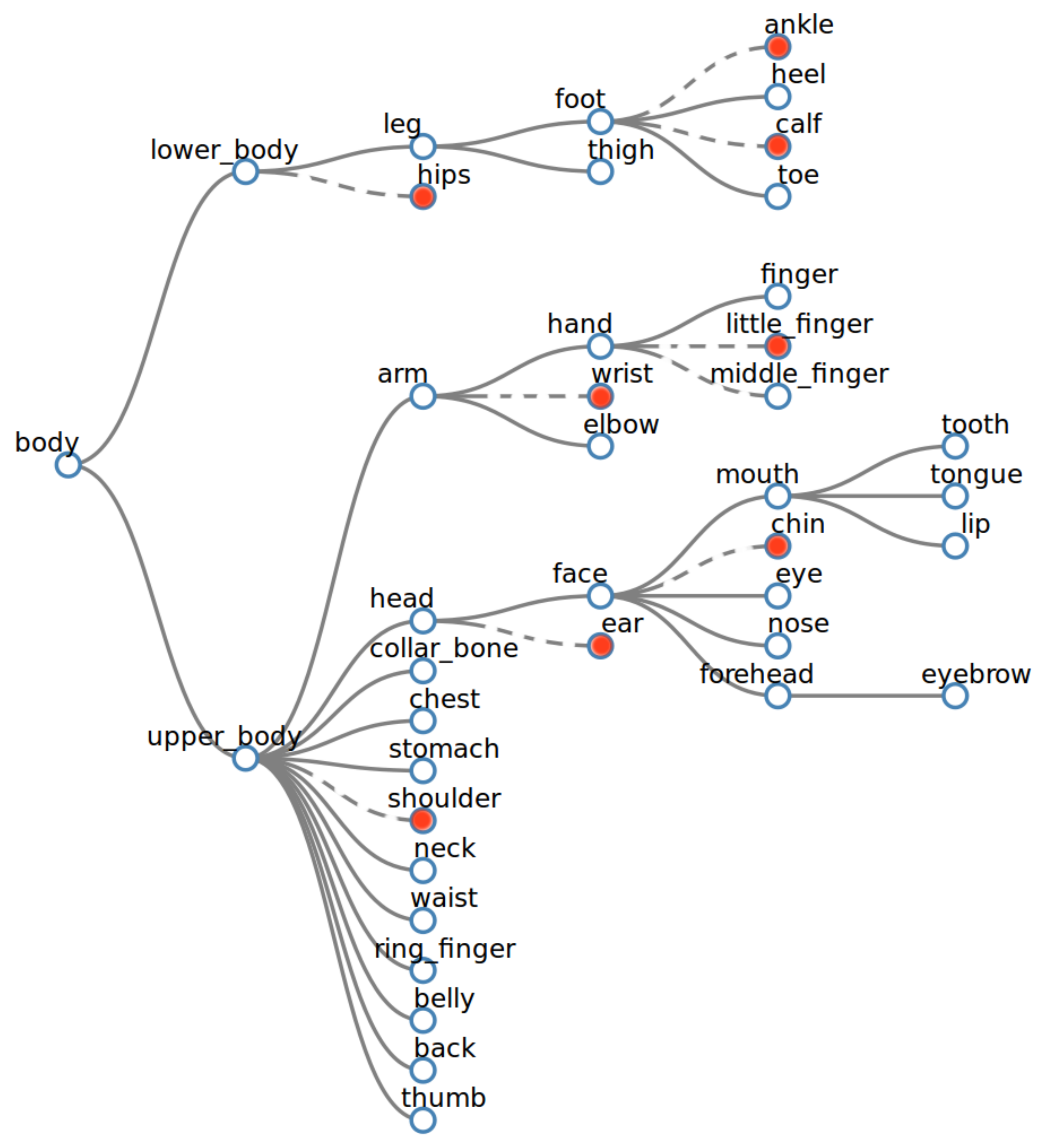}}\hfill
\fbox{\includegraphics[height=29ex,width=40ex]{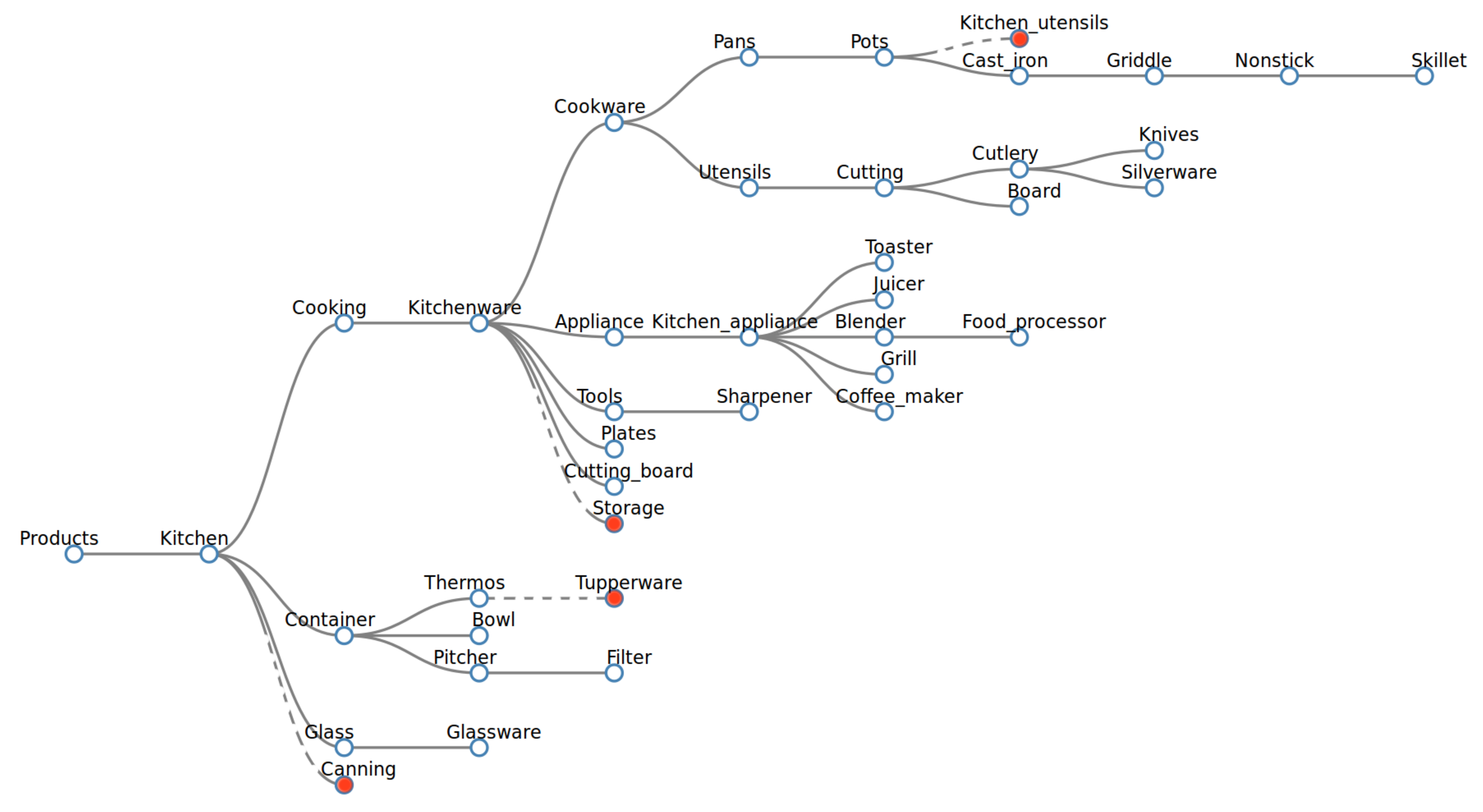}}\hfill
\fbox{\includegraphics[height=29ex,width=40ex]{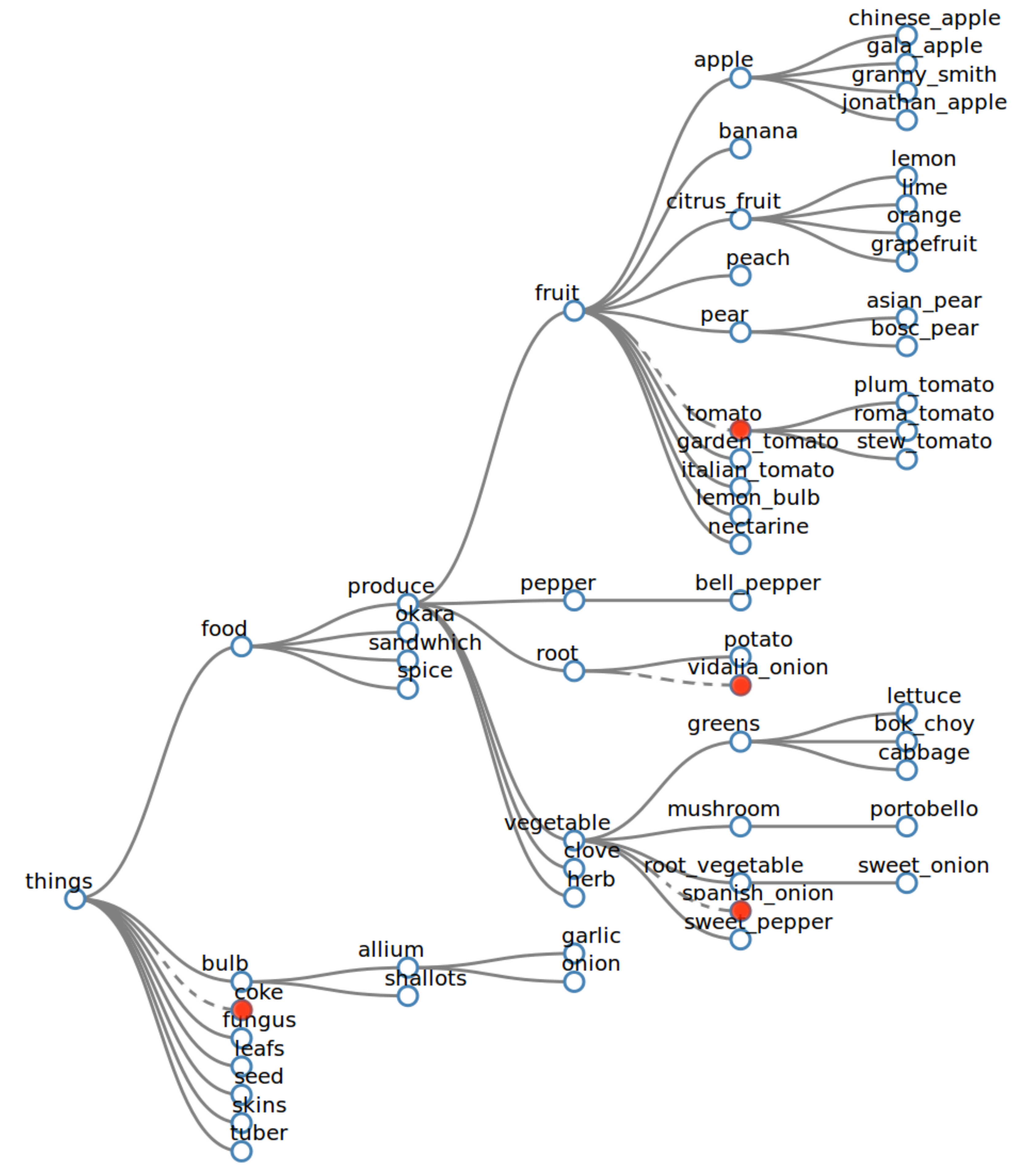}}
\caption{\small{MAP hierarchies representing body parts, amazon kitchen products, and food items
    (left to right). Red nodes indicate items for which the parent edge has high uncertainty (marginal probability below 0.75). Videos showing the whole process of hierarchy generation can be found on our project page: \texttt{http://rse-lab.cs.washington.edu/projects/learn-taxonomies}.}}  \label{fig:hierarchies} \end{center} \end{figure*}

In these experiments we provide examples demonstrating that our method can be applied to different tasks using AMT. The following pipeline is followed for all three application domains: collect the set of concepts; design the ``path question'' to ask; collect multiple answers for all possible ``path questions''; estimate hierarchies using our approach. Collecting answers for all possible questions enabled us to test different settings and methods without collecting new data for each experiment. AMT is used to gather answers for ``path questions''. The process for different domains is almost identical: We ask workers to answer ``true-or-false'' questions regarding a path in a hierarchy. Our method is able to consider the noise rate of workers. We estimate this by gathering answers from 8 workers for each question, then take the majority vote as the answer, and use all answers to determine the noise ratio for that question. Note that noise ratios not only capture the inherent noise in using AMT, but also the uncertainty of people about the relationship between concepts. 5 different path questions are put into one Human Intelligence Task (HIT). Each HIT costs \$0.04. The average time for a worker to finish one HIT is about 4 seconds.  

The process of building the hierarchies is divided into two consecutive phases. In the first phase, a distribution is built using a subset of the concepts. In the second phase, we use the process of inserting new concepts into the hierarchy, until all concepts are represented. For the body part dataset, we randomly chose 10 concepts belonging to the first phase. For online Amazon shopping and RGBD object data, the initial set is decided by thresholding the frequency of the words used by workers to tag images (15 nodes for Amazon objects and 23 nodes for RGBD objects).  The learned MAP hierarchies are shown in Fig.~\ref{fig:hierarchies}.  




\subsubsection{Representing Body Parts}

\noindent Here, we want to build a hierarchy to visualize the ``is a part of'' relationship between body parts. The set of body part words are collected using Google search. An example path question would be ``Is \textit{ear} part of \textit{upper body}?''.  The MAP tree after asking 2,000 questions is shown in the left panel of Fig.~\ref{fig:hierarchies}. As can be seen, the overall structure agrees very well with people's common sense of the human body structure. Some of the nodes in the tree are shown in red, indicating edges whose marginal probability is below $0.75$.  These edges also reflect people's uncertainty in the concept hierarchy. For example, it is not obvious whether \textit{ear} should be part of the \textit{head} or \textit{face}, the second most likely placement. Similarly, it is not clear for people whether \textit{ankle} should be part of \textit{foot} or \emph{leg}, and whether \textit{wrist} should be part of \emph{arm} or \emph{hand}.

An obvious mistake made by the system is that \textit{ring finger} and \textit{thumb} are connected to the \textit{upper body} rather than \textit{hand}. This is caused by questions such as ``Is \textit{ring finger} part of \textit{arm}?'', which only 1 out of 8 workers answered with yes. Hence the concept of \textit{ring finger} or \textit{thumb} is not placed into a position below \textit{arm}.

\subsubsection{Online Shopping Catalogue}

The second task is to arrange kitchen products taken from the Amazon website. There are some existing kitchen hierarchies, for example, the Amazon hierarchy~\cite{amazon:2015}. However, the words used by Amazon, for example, ``Tools-and-Gadgets'', might be quite confusing for customers. Therefore, we collected the set of words used by workers in searching for products. We provide AMT workers images of products, and ask them to write down words they would like to see in navigating kitchen products. Some basic preprocessing is done to merge plural and singular of the same words, remove obviously wrong words, and remove tags used less than 5 times by workers because they might be some nicknames used by a particular person. We also remove the word ``set'', because it is used by workers to refer to a ``collection of things'' (\eg, pots set, knives set), but not related to the type of products shown in the pictures. The path questions have the form ``Would you try to find \textit{pots} under the category of \textit{kitchenware}?''  The learned MAP tree is shown in the middle panel of Fig.~\ref{fig:hierarchies}. 


\subsubsection{Food Item Names}

This experiment investigates learning a hierarchy over food items used in a robotics setting~\cite{Lai:icra2011}, where the goal is to learn names people use in a natural setting to refer to objects.  Here, AMT workers were shown images from the RGBD object dataset~\cite{Lai:icra2011} and asked to provide names they would use to refer to these objects. Some basic pre-processing was done to remove noisy tags and highly infrequent words. The path questions for this domain are of the form ``Is it correct to say all \textit{apples} are \textit{fruits}?''.

The MAP tree is shown in the right panel of Fig.~\ref{fig:hierarchies}.  Again, while the tree captures the correct hierarchy mostly, high uncertainty items provide interesting insights. For instance, \textit{tomato} is classified as \textit{fruit} in the MAP tree, but also has a significant probability of being a \emph{vegetable}, indicating people's uncertainty, or disagreement, about this concept.  Meanwhile, the crowd of workers was able to uncover very non-obvious relationships such as \textit{allium} is a kind of \textit{bulb}.



\section{Conclusion}
\label{sec:conclusion}

\looseness -1 We introduced an approach for learning hierarchies over concepts using crowdsourcing.  Our approach incorporates simple questions that can be answered by non-experts without global knowledge of the concept domain.  To deal with the inherent noise in crowdsourced information and with people's uncertainty, and possible disagreement, about hierarchical relationships, we develop a Bayesian framework for estimating posterior distributions over hierarchies.  When new answers become available, these distributions are updated efficiently using a sampling-based approximation for the intractably large set of possible hierarchies.  The Bayesian treatment also allows us to actively generate queries that are most informative given the current uncertainty. New concepts can be added to the hierarchy at any point in time, automatically triggering queries that enable the correct placement of these concepts.  It should also be noted that our approach lends itself naturally to manual correction of errors in an estimated hierarchy: by setting the weights inconsistent with a manual annotation to zero, the posterior over trees automatically adjusts to respect this constraint.

We investigated several aspects of our framework and demonstrated that it is able to recover quite good hierarchies for real world concepts using AMT. Importantly, by reasoning about uncertainty over hierarchies, our approach is able to unveil confusion of non-experts over concepts, such as whether tomato is a fruit or vegetable, or whether the wrist is part of a person's arm or hand.  
\looseness -1 We believe that these abilities are extremely useful for applications where hierarchies should reflect the knowledge or expectations of regular users, rather than domain experts.  Example applications could be search engines for products or restaurants, or robots interacting with people who use various terms to relate to objects in the world.  Investigating such use cases is an interesting avenue for future research.  Other possible future directions include explicit treatment of synonyms and automatic detection of inconsistencies.  

\section*{Acknowledgement}

This work was funded in part by the Intel Science and Technology Center for Pervasive Computing, ARO grant W911NF-12-1-0197, ERC StG 307036 and the Nano-Tera.ch program as part of the Opensense II project. We would like to thank Jonathan Bragg for generously sharing the data and code. We would also like to thank Tianyi Zhou for helpful discussions.


\bibliographystyle{named}
\small
\bibliography{ijcai15_crowd}
\clearpage
\onecolumn
\section{Derivation of the Objective Function}

\begin{align}
&L^{\beta}_{\tilde{\pi}}(\Lambda') - L^{\beta}_{\tilde{\pi}}(\Lambda) \\
=& \sum_{T\in \mathcal{T}} \tilde{\pi}(T) \log P(T|\Lambda) - \sum_{T\in \mathcal{T}} \tilde{\pi}(T) \log P(T|\Lambda') + \sum_{i,j} \beta (|\lambda_{i,j}'| - |\lambda_{i,j}|) \label{eq:ori}\\
=& \sum_{T\in \mathcal{T}} \tilde{\pi}(T) \log \frac{\exp(\sum_{e_{i,j}\in T}\lambda_{i,j}) }{\exp(\sum_{e_{i,j}\in T}\lambda_{i,j} + \delta_{i,j})} + \sum_{T\in \mathcal{T}} \tilde{\pi}(T)\log \frac{Z(\Lambda')}{Z(\Lambda)} + \sum_{i,j} \beta(|\lambda_{i,j}' - |\lambda_{i,j}|)\label{eq:rewrite}\\
=&\sum_{T\in \mathcal{T}}\tilde{\pi}(T)\sum_{e_{i,j}\in T} -\delta_{i,j} + \log \frac{Z(\Lambda')}{Z(\Lambda')} + \sum_{i,j} \beta(|\lambda_{i,j}'| - |\lambda_{i,j}|) \label{eq:rewrite_delta}\\
=& \sum_{i,j} -\delta_{i,j} \tilde{P}(e_{i,j}) + \log \frac{\sum_{T'\in \mathcal{T}} \exp(\sum_{e_{i,j}\in T'} \lambda_{i,j} + \delta_{i,j})}{\sum_{T\in \mathcal{T}} \exp(\sum_{e_{i,j}\in T}\lambda_{i,j})} + \sum_{i, j} \beta(|\lambda_{i,j}'| - |\lambda_{i,j}|)  \\
=& \sum_{i,j} -\delta_{i,j}\tilde{P}(e_{i,j}) + \log \sum_{T\in \mathcal{T}} P(T|\Lambda) \exp(\sum_{e_{i,j}\in T}\delta_{i,j}) + \sum_{i, j} \beta(|\lambda_{i,j}'| - |\lambda_{i,j}|) \\
=& \sum_{i, j} -\delta_{i,j} \tilde{P}(e_{i,j}) + \log \sum_{T\in\mathcal{T}} P(T|\Lambda) \exp(\sum_{e_{i,j}\in T} \frac{1}{N} N\delta_{i,j}) + \sum_{i,j}\beta(|\lambda_{i,j}'| - |\lambda_{i,j}|)\label{eq:ineq_left}\\
\leq&\sum_{i,j} -\delta_{i,j} \tilde{P}(e_{i,j}) + \frac{1}{N}P(e_{i,j})(e^{N\delta_{i,j}} - 1) + \beta(|\lambda_{i,j}'| - |\lambda_{i,j}|)\label{eq:upper_bound}
\end{align}

(\ref{eq:ori}) $\to$ (\ref{eq:rewrite}) uses the definition of $P(T|\Lambda)$.

(\ref{eq:rewrite}) $\to$ (\ref{eq:rewrite_delta}) uses the fact that

\[\sum_{T\in \mathcal{T}} \tilde{\pi}(T)\sum_{e_{i,j}\in T} -\delta_{i,j} = \sum_{{i,j}}-\delta_{i,j}\sum_{T\in \mathcal{T}:e_{i,j}\in T} \tilde{\pi}(T) = \sum_{{i,j}}-\delta_{i,j}\tilde{P}(e_{i,j}),\]
where $\tilde{P}(e_{i,j})$ is the marginal likelihood of the edge $e_{i,j}$.

To get (\ref{eq:ineq_left}) $\to$ (\ref{eq:upper_bound}), we use an inequality that, if $x_j \in \mathbb{R}$ and $p_j \geq 0$ with $\sum_{j}p_j \leq 1$, then

\[\exp(\sum_{j} p_j x_j) - 1 \leq \sum_{j}p_j(e^{x_j} - 1).\]

Such that 

\[\exp(\sum_{e_{i,j}\in T}\frac{1}{N} N\delta_{i,j})\leq 1 + \sum_{e_{i,j}\in T}\frac{1}{N}(e^{N\delta_{i,j}} - 1).\]

The it follows

\begin{align}
&\log \sum_{T\in \mathcal{T}} P(T|\Lambda)\exp(\sum_{e_{i,j}\in T}\frac{1}{N}N\delta_{i,j})\\
\leq& \log(1 + \sum_{T\in \mathcal{T}}P(T|\Lambda)\sum_{e_{i,j}\in T}\frac{1}{N}(e^{N\delta_{i,j}} - 1))\\
=&\log(1 + \frac{1}{N}\sum_{i,j}P(e_{i,j})(e^{N\delta_{i,j}} - 1))\\
\leq&\frac{1}{N}\sum_{i,j}P(e_{i,j})(e^{N\delta_{i,j}} - 1)\\
\end{align}

(11) $\to$ (12) is true because $\log(1+x) \leq x, \forall x\geq -1$, and 

\[\frac{1}{N}\sum_{i,j}P(e_{i,j}) (e^{N\delta_{i,j}} - 1)\geq \frac{1}{N}\sum_{i,j}-P(e_{i,j}) = -1.\]

\section{Minimization of (\ref{eq:upper_bound})}

Case 1: $\delta_{i,j} = -\lambda_{i,j}$. Such that (\ref{eq:upper_bound}) becomes

\begin{equation}
\sum_{i,j} \lambda_{i,j} \tilde{P}(e_{i,j}) + \frac{1}{N} P(e_{i,j}) (e^{-N\lambda_{i,j}} - 1)\label{eq:case1}
\end{equation}

Case 2: $\lambda_{i,j} + \delta_{i,j} \geq 0$. Such that (\ref{eq:upper_bound}) becomes

\begin{equation}
\sum_{i,j} -\delta_{i,j} \tilde{P}(e_{i,j}) + \frac{1}{N} P(e_{i,j}) (e^{N\delta_{i,j}} - 1) + \beta \delta_{i,j}\label{eq:case2}
\end{equation}

Take derivative of (\ref{eq:case2}), and set it to be 0:

\[-\tilde{P}(e_{i,j}) + P(e_{i,j})e^{N\delta_{i,j}} + \beta = 0,\]
the solution is 

\[\frac{1}{N} \log \frac{\tilde{P}(e_{i,j}) - \beta}{P(e_{i,j})}.\]

Case 3: $\lambda_{i,j} + \delta_{i,j} < 0$. Such that (\ref{eq:upper_bound}) becomes

\begin{equation}
\sum_{i,j} -\delta_{i,j} \tilde{P}(e_{i,j}) + \frac{1}{N} P(e_{i,j}) (e^{N\delta_{i,j}} - 1) - \beta \delta_{i,j}\label{eq:case3}
\end{equation}

Take derivative of (\ref{eq:case3}), and set it to be 0:

\[-\tilde{P}(e_{i,j}) + P(e_{i,j})e^{N\delta_{i,j}} - \beta = 0,\]
the solution is 

\[\frac{1}{N} \log \frac{\tilde{P}(e_{i,j}) + \beta}{P(e_{i,j})}.\]

\section{Proof of Theorem 1}

\begin{theorem}
Assume $\beta$ is strictly positive. Then  Algorithm 1 produces a sequences $\Lambda^{(1)}, \Lambda^{(2)}, \ldots$ such that
\[\lim_{\ell \rightarrow \infty} L_{\tilde{\pi}}^{\beta}(\Lambda^{(\ell)}) = \min_{\Lambda} L_{\tilde{\pi}}^{\beta}(\Lambda).\]
\vspace{-1ex}
\end{theorem}

\begin{proof}
First let us define $\Lambda^{+}$ and $\Lambda^{-}$ in terms of $\Lambda$ as follows: for each $(i,j)$, if $\lambda_{i,j} \geq 0$, then $\lambda_{i,j}^{+} = \lambda_{i,j}$ and $\lambda_{i,j}^{-} = 0$, and if $\lambda_{i,j}\leq 0$, then $\lambda_{i,j}^{+} = 0$ and $\lambda_{i,j}^{-} = -\lambda_{i,j}$. $\Lambda'^{+}$, $\Lambda'^{-}$, $\Lambda^{(\ell)+}$, $\Lambda^{(\ell)-}$, etc. are defined analogously.

Let $F_{i,j}$ denote the $(i,j)$ component in ~(\ref{eq:upper_bound}), For any $\Lambda$ and $\Delta$, we have the following:

\begin{align}
|\lambda + \delta| - |\lambda| = \min \{\delta^{+} + \delta^{-}|\delta^{+} \geq -\lambda^{+}, \delta^{-} \geq -\lambda^{-}, \delta^{+} - \delta^{-} = \delta\}
\end{align}
Plugging into the definition of $F_{i,j}$ gives:

\begin{align}
F_{i,j}(\Lambda, \Delta) =& -\delta_{i,j} \tilde{P}(e_{i,j}) + \frac{1}{N}P(e_{i,j})(e^{N\delta_{i,j}} - 1) + \beta(|\lambda_{i,j} + \delta_{i,j}| - |\lambda_{i,j}|)\\
=& \min\{G_{i,j}(\Lambda, \Delta^{+}, \Delta^{-})|\delta_{i,j}^{+} \geq -\lambda_{i,j}^{+}, \delta_{i,j}^{-}\geq \lambda_{i,j}^{-}, \delta_{i,j}^{+}-\delta_{i,j}^{-} = \delta_{i,j}\},
\end{align}
where

\begin{align}
G_{i,j}(\Lambda, \Delta^{+}, \Delta^{-}) = (\delta_{i,j}^{-} - \delta_{i,j}^{+})\tilde{P}(e_{i,j}) + \frac{1}{N}P(e_{i,j})(e^{N(\delta_{i,j}^{+} - \delta_{i,j}^{-})} - 1) + \beta (\delta_{i,j}^{+}+\delta_{i,j}^{-})
\end{align}

So, by~(\ref{eq:upper_bound}),

\begin{align}
L^{\beta}_{\tilde{\pi}}(\Lambda^{(\ell+1)}) - L^{\beta}_{\tilde{\pi}}(\Lambda^{(\ell)}) \leq & \sum_{i,j}F_{i,j}(\Lambda^{\ell}, \Delta)\\
= & \sum_{i,j} \min_{\delta_{i,j}} F_{i,j}(\Lambda^{\ell}, \Delta)\\
=& \sum_{i,j} \min\{G_{i,j}(\Lambda^{\ell}, \Delta^{+}, \Delta^{-})|\delta_{i,j}^{+} \geq -\lambda_{i,j}^{+}, \delta_{i,j}^{-}\geq \lambda_{i,j}^{-}, \delta_{i,j}^{+}-\delta_{i,j}^{-} = \delta_{i,j}\}\}
\end{align}

Note that $G_{i,j}(\Lambda, \bm{0}, \bm{0}) = 0$, so none of the terms in this sum can be positive. So the $\Lambda^{\ell}$'s have a convergent subsequence converging to some $\hat{\Lambda}$ such that

\begin{align}\label{eq:convergence}
\sum_{i,j} \min\{G_{i,j}(\Lambda^{\ell}, \Delta^{+}, \Delta^{-})|\delta_{i,j}^{+} \geq -\lambda_{i,j}^{+}, \delta_{i,j}^{-}\geq \lambda_{i,j}^{-}, \delta_{i,j}^{+}-\delta_{i,j}^{-} = \delta_{i,j}\}\} = 0.
\end{align}

It is easy to verify that minimizing $L_{\tilde{\pi}}^{\beta}(\Lambda)$ is the dual problem of the following convex program:

\begin{align}
&\max_{p_1, \ldots, p_N \in \mathbb{R}^{+N}} \sum_{i = 1}^{N} H(p_{i})\\
s.t. & \sum_{i = 0}^{N} p(e_{i,j}) = 1, \forall j\\
& \tilde{P}(e_{i,j}) - P(e_{i,j}) \leq \beta, \forall (i,j)\\
& P(e_{i,j}) - \tilde{P}(e_{i,j}) \leq \beta, \forall (i,j).
\end{align}

We will show that $\hat{\Lambda}^{+}$ and $\hat{\Lambda}^{-}$ together with $P(T|\hat{\Lambda})$ satisfy the KKT condition of the previous convex program, and thus form a solution to the prime problem as well as to the dual, the minimization of $L_{\tilde{\pi}}^{\beta}$. For $P(T|\hat{\Lambda})$, these conditions work out to be the following for all $(i,j)$:

\begin{align}
\hat{\lambda}_{i,j}^{+} \geq 0, \tilde{P}(e_{i,j}) - P(e_{i,j}) \leq \beta, \hat{\lambda}_{i,j}^{+}(\tilde{P}(e_{i,j}) - P(e_{i,j}) - \beta) = 0\label{eq:kkt1}\\
\hat{\lambda}_{i,j}^{-} \geq 0, P(e_{i,j}) - \tilde{P}(e_{i,j})  \leq \beta, \hat{\lambda}_{i,j}^{+}(P(e_{i,j}) - \tilde{P}(e_{i,j}) - \beta) = 0\label{eq:kkt2}
\end{align}

Since $G_{i,j}(\hat{\Lambda}, \bm{0}, \bm{0}) = 0$, by~(\ref{eq:convergence}), if $\hat{\lambda}_{i,j} >0$ then $G_{i,j}(\Lambda, \Delta^{+}, \bm{0})$ is nonnegative in a neighborhood of $\delta_{i,j}^{+} = 0$, and so has a local minimum at this point. Such that

\begin{align}
\frac{\partial G_{i,j}(\Lambda, \Delta^{+}, \bm{0})}{\partial \delta_{i,j}^{+}}|_{\delta_{i,j}^{+} = 0} = -\tilde{P}(e_{i,j}) + P(e_{i,j}) + \beta = 0.
\end{align}

If $\hat{\lambda}^{+}_{i,j} = 0$, then~(\ref{eq:convergence}) gives that $G_{i,j}(\hat{\Lambda}, \bm{0}, \bm{0}) = 0$ for $\delta^{+}_{i,j} \geq 0$. Thus $\partial G_{i,j}(\Lambda, \Delta^{+}, \bm{0})$ cannot be decreasing at $\delta^{+}_{i,j} = 0$. Therefore, the partial derivative above must be nonnegative. Altogether, these prove~(\ref{eq:kkt1}). (\ref{eq:kkt2}) can be proved analogously.

As a whole, we proved that

\[\lim_{\ell \rightarrow \infty} L_{\tilde{\pi}}^{\beta}(\Lambda^{(\ell)}) = L_{\tilde{\pi}}^{\beta}(\hat{\Lambda}) = \min_{\Lambda} L_{\tilde{\pi}}^{\beta}(\Lambda).\]

\end{proof}

\section{Proof of Theorem 2}

\begin{lemma}\label{lemma:1}
Suppose samples $\tilde{\pi}$ are obtained from any tree distribution $\pi$. Then
\[|L_{\tilde{\pi}}(\Lambda) - L_{\pi}(\Lambda)| \leq \sum_{i = 0}^{N}\sum_{j=1}^{N}|\lambda_{i,j}||\tilde{P}(e_{i,j}) - P(e_{i,j})|,\]
where $P(e_{i,j}) = \sum_{T\in \mathcal{T}:e_{i,j}\in T} \pi(T)$ and $\tilde{P}(e_{i,j}) = \sum_{T\in \mathcal{T}:e_{i,j}\in T} \tilde{\pi}(T)$.
\end{lemma}
\begin{proof}
\begin{align}
L_{\tilde{\pi}}(\Lambda) = \sum_{T\in \mathcal{T}} \tilde{\pi}(T) \log \frac{\exp{\sum_{e_{i,j}\in T} \lambda_{i,j}}}{Z(\Lambda)}
\end{align}
Such that
\begin{align}
|L_{\tilde{\pi}}(\Lambda) - L_{\pi}(\Lambda)| = |\sum_{i,j} \lambda_{i,j}(\tilde{P}(e_{i,j}) - P(e_{i,j}))|\leq \sum_{i=0}^{N}\sum_{j=1}^{N} |\lambda_{i, j}||\tilde{P}(e_{i,j}) - P(e_{i,j})|.
\end{align}
\end{proof}

\begin{lemma}\label{lemma:2}
Suppose samples $\tilde{\pi}$ are obtained from any tree distribution $\pi$. Assume that $|P(e_{i,j}) - \tilde{P}(e_{i,j})| \leq \beta_{i,j}, \forall (i,j)$, Let $\hat{\Lambda}$ minimize the regularized log loss $L^{\beta}_{\tilde{\pi}}(\Lambda)$. The for every $\Lambda$ it holds that
\[L_{\pi}(\hat{\Lambda}) \leq L_{\pi}(\Lambda) + 2\sum_{i=0}^N\sum_{j=1}^N \beta |\lambda_{i,j}|.\]
\end{lemma}
\begin{proof}
\begin{align}
L_{\pi}(\hat{\Lambda}) \leq & L_{\tilde{\pi}}(\hat{\Lambda}) + \sum_{i,j} \beta |\hat{\lambda}_{i,j}| = L_{\tilde{\pi}}^{\beta}(\hat{\Lambda})\label{eq:lemma2_1}\\
 \leq & L_{\tilde{\pi}}^{\beta}(\Lambda) = L_{\tilde{\pi}}(\Lambda) + \sum_{i,j}\beta |\lambda_{i,j}|\label{eq:lemma2_2}\\
 \leq & L_{\pi}(\Lambda) + 2\sum_{i,j} \beta |\lambda_{i,j}|.\label{eq:lemma2_3}
\end{align}
\end{proof}

(~\ref{eq:lemma2_1}) to (~\ref{eq:lemma2_2}) is tree because of the optimality of $\hat{\Lambda}$. (~\ref{eq:lemma2_2}) to (~\ref{eq:lemma2_3}) follow from Lemma~\ref{lemma:1}.

\begin{theorem}\label{theorem:loss_bound}
Suppose $m$ samples $\tilde{\pi}$ are obtained from any tree distribution $\pi$.
Let $\hat{\Lambda}$ minimize the regularized log loss $L_{\tilde{\pi}}^{{\beta}}(\Lambda)$ with $\beta = \sqrt{\log(N/\delta)/m}$. Then for every $\Lambda$ 
it holds with probability at least $1-\delta$ that
\[L_{\pi}(\hat{\Lambda}) \leq L_{\pi}(\Lambda) + 2\|\Lambda\|_1\sqrt{\log(N/\delta)/m}\]
\end{theorem}
\begin{proof}
By Hoeffding's inequality, for a fixed pair of $(i,j)$, the probability that $P(e_{i,j}) - \tilde{P}(e_{i,j})$ exceeds $\beta$ is at most $e^{-2\beta^2m} = \frac{\delta}{N^2}$. By the union bound, the probability of this happening for any pair of $(i,j)$ is at most $\delta$. Then the theorem follows from Lemma~\ref{lemma:2}.
\end{proof}

\end{document}